\documentclass{article}
\usepackage{ijcai17}

\usepackage{times}

\usepackage{CJKutf8}
\usepackage{amsmath}
\usepackage{graphicx}
\usepackage{multirow}
\usepackage[linesnumbered,ruled,vlined]{algorithm2e}
\usepackage{url}

\usepackage{amsfonts}
\usepackage{colortbl}

\definecolor{mygray}{gray}{.9}
\definecolor{mypink}{rgb}{.99,.91,.95}
\definecolor{mycyan}{cmyk}{.3,0,0,0}

\newcommand{\newcite}[1]{\citeauthor{#1}~\shortcite{#1}}

\title{A Deep Neural Network for Chinese Zero Pronoun Resolution}

\author{Qingyu Yin, Weinan Zhang, Yu Zhang, Ting Liu\thanks{Email corresponding.} \\
  Research Center for Social Computing and Information Retrieval \\
  Harbin Institute of Technology, China \\
  {\{qyyin, wnzhang, yzhang, tliu\}@ir.hit.edu.cn} \\
  }

\begin{document}
\begin{CJK*}{UTF8}{gbsn}
\maketitle
\begin{abstract}
Existing approaches for Chinese zero pronoun resolution overlook semantic information. This is because zero pronouns have no descriptive information, which results in difficulty in explicitly capturing their semantic similarities with antecedents. Moreover, when dealing with candidate antecedents, traditional systems simply take advantage of the local information of a single candidate antecedent while failing to consider the underlying information provided by the other candidates from a global perspective. To address these weaknesses, we propose a novel zero pronoun-specific neural network, which is capable of representing zero pronouns by utilizing the contextual information at the semantic level. In addition, when dealing with candidate antecedents, a two-level candidate encoder is employed to explicitly capture both the local and global information of candidate antecedents. We conduct experiments on the Chinese portion of the OntoNotes 5.0 corpus. Experimental results show that our approach substantially outperforms the state-of-the-art method in various experimental settings.
\end{abstract}

\section{Introduction}
Zero pronoun (ZP) refers to the component that is omitted due to the coherence of language. A ZP can be either anaphoric if it corefers to one or more noun phrases (NPs) in the preceding text, or non-anaphoric if there are no such NPs. These NPs that provide the necessary information for interpreting the gaps are normally described as antecedents.
Following shows an example of ZPs and their antecedents, where ``$\phi$'' is used to denote a ZP.
        \begin{quote}\small
            [警方] 怀疑\ 这是\ 一起\ 黑枪\ 案件， {\bf $\phi_1$ }\ 将\ 枪械\ 和\ 皮包\ 交送\ 市里\ {\bf $\phi_2$}\ 以\ 清理\ 案情。

            ([The police] suspected that this is a criminal case about illegal guns, {\bf $\phi_1$ } brought the guns and bags to the city {\bf $\phi_2$ }  to deal with the case.)
        \end{quote}
    The ZP ``$\phi_1$'' in this example is an anaphoric zero pronoun (AZP) that refers to the NP ``警方/The police'' while another ZP ``$\phi_2$'' is non-anaphoric. 

ZP is ubiquitous in the pro-drop languages, such as Chinese, Japanese, etc.  According to~\cite{kim2000}, the use of overt subjects in English is over 96\% while the percentage is only 64\% in Chinese, indicating that the ZP phenomenon in Chinese is much more prevalent. Meanwhile, important grammatical attributes such as gender and number that have been proven to be essential in pronoun resolution \cite{chen2014chinese}, is absent to ZPs. Therefore, it is challenging but also crucial to resolving ZPs, especially for the pro-drop languages such as Chinese. 

Existing researches on Chinese ZP resolution can be classified into two types: supervised approaches \cite{zhao2007,kong2010,chen2013,chen2016} and unsupervised ones \cite{chen2014,chen2015}. In these studies, ZP resolution is typically composed of two steps, i.e., anaphoric zero pronoun (AZP) identification that identifies whether a given ZP is anaphoric; and AZP resolution, which determines the exact antecedent for an AZP.

For AZP resolution, existing supervised approaches \cite{zhao2007,kong2010,chen2013,chen2016} typically utilize only syntactic and lexical features while ignoring semantic information that plays an important role in the resolution of common NPs~\cite{ng2007semantic}. This is because ZPs have no descriptive information, which makes it almost impossible to calculate semantic similarities and relatedness scores between the ZP and its candidate antecedents. One straightforward way to address this issue is to represent ZPs with some available information. Inspired by \newcite{chen2016} who utilize a ZP's governing verb and preceding word to encode its lexical contexts, we notice that a ZP's contexts can help to describe the ZP itself. For example, given a sentence ``$\phi$ 很甜\ /$\phi$ tastes sweet'', people may resolve the ZP ``{\bf $\phi$ }'' to the candidate antecedent  ``an apple'', but will never regard ``a book'' as its antecedent, because they naturally look at the ZP's context ``tastes sweet'' to resolve it (``a book'' cannot be ``sweet''). These motivate us to seek an effective method to represent ZPs by utilizing contextual information at the semantic level.

In addition, when dealing with candidate antecedents, previous researches \cite{zhao2007,chen2015,chen2016} operate solely by extracting features on one single candidate at a time. However, treating candidate antecedents in isolation is restricted to modeling their semantic level information, because such a strategy simply takes advantage of the local information of a single candidate antecedent while failing to consider the information provided by the other candidates from a global perspective. We argue that the incorporation of global information allows the model to capture the underlying information of antecedents and thereby benefits the resolution of ZPs. As an example for its usefulness, it might be ambiguous whether the candidate antecedent ``它/it'' corefers to the ZP ``$\phi$'' in the sentence ``我 喜欢 读 $\phi$ /I like to read $\phi$'' if we consider its local information solely.
However, it is easy to infer that ``它/it'' could hardly be the correct antecedent if we take advantage of the global information obtained from the candidate set $\{$``一些书/some books'', ``一个书包/a bag'', ``它/it'' $\}$, because ``它/it'' possibly contains the similar information as ``一个书包/a bag'' and ``$\phi$'' requires some reading matters to fill in the gap. Therefore, to obtain the comprehensive representation of a given candidate antecedent, a desirable model should be capable of having access to the useful information of other candidate antecedents from a global perspective, and it is precisely this access to the global information that local models lack.

To address the above weaknesses, in this paper, we focus on AZP resolution, proposing a deep neural network to perform the task.  In particular, we develop the zero pronoun-specific neural network ({\bf ZPSNN}), which we believe has the following advantages.

First, we make an extension on traditional long short-term memory (LSTM) network, building the ZP-centered LSTM that is competent for representing a ZP by modeling its preceding and following contexts.
Second, to capture both the local and global information of candidate antecedents, we propose a novel two-level candidate antecedent encoder, which is consists of two neural network components, i.e., the $local\ encoder$ that generates the local representation of a candidate antecedent by utilizing its context and content; and the $global\ encoder$ that encodes a candidate antecedent with the useful information provided by the other candidates from the global perspective. We believe our candidate antecedent encoder is capable of generating comprehensive representations of candidate antecedents by incorporating sufficient local and global information, and thereby provides reliable hints for inferring the antecedents.
Using these continuous distributed representations, our resolver takes advantage of semantic information when resolving ZPs. Furthermore, since that every component of {\bf ZPSNN} is differentiable, the entire model could be efficiently trained end-to-end with gradient descent, where the loss function is the cross-entropy error of coreference classification. We conduct experiments on the Chinese portion of the OntoNotes 5.0 corpus, comparing with the baseline system in different experimental settings. Experimental results show that our method yields the state-of-the-art performance.

The major contributions of the work presented in this paper are as follows.
\begin{itemize}
\item We develop a zero pronoun-specific neural network ({\bf ZPSNN}) for Chinese ZP resolution. 
Our approach is data-driven and could be trained in an end-to-end way with standard backpropagation.
\item Our model is capable of capturing the local and global information of candidate antecedents by employing an encoder with two-level architecture. Meanwhile, we encode the ZPs into vector representations by utilizing their contextual information. Using these continuous distributed representations, our system takes advantage of semantic information when resolving ZPs.
\item We empirically verify that our model outperforms the state-of-the-art approach on the Chinese portion of the OntoNotes 5.0 corpus.
\end{itemize}

        \begin{figure*}
            \centering
            \includegraphics[width=0.95\textwidth]{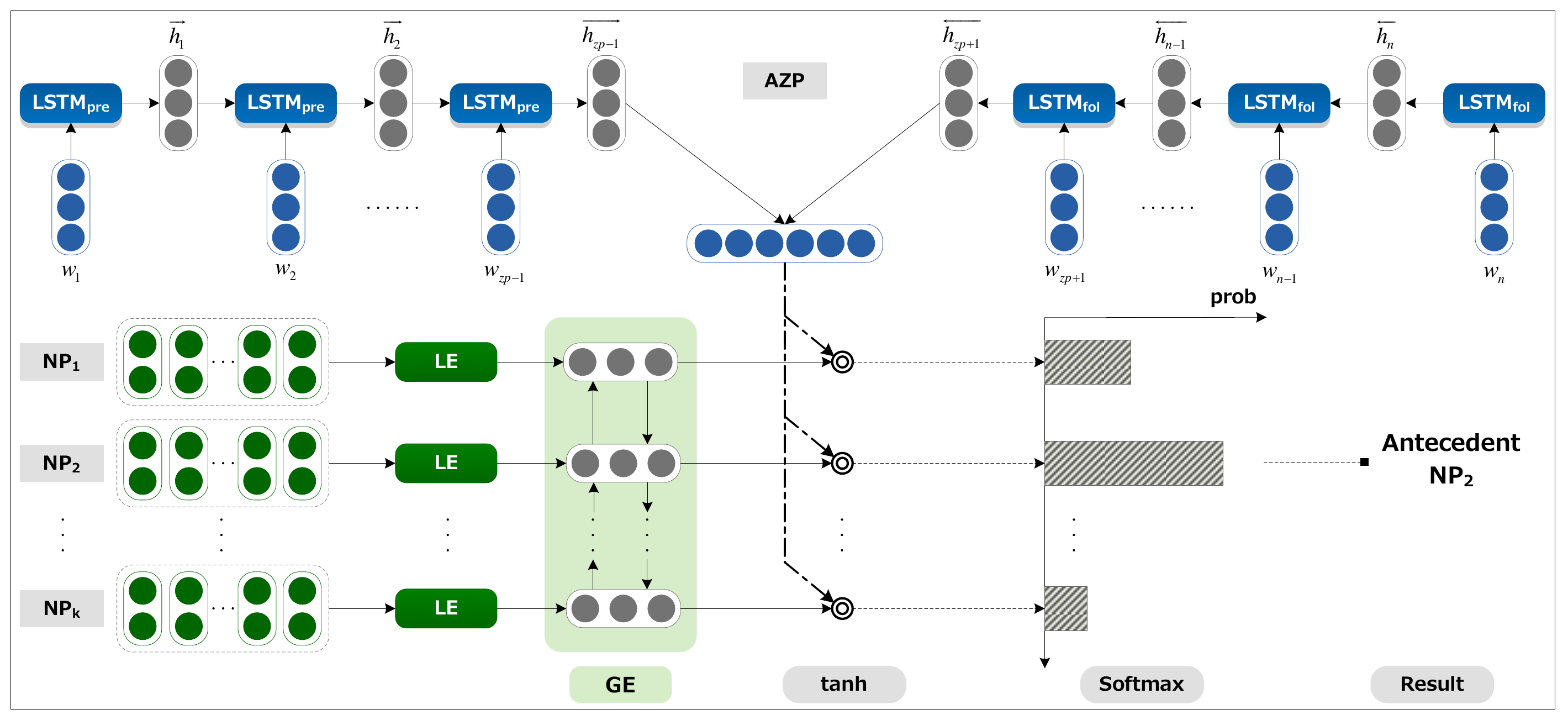}
            \caption{The zero pronoun-specific neural network ({\bf ZPSNN}) for AZP resolution. For AZP part, $w_{i}$ means the $i$-th word in the sentence, $w_{zp-i}$ is the $i$-th last word before the AZP and $w_{zp+i}$ is the $i$-th word behind the AZP. For candidate antecedent part, {\bf LE} and {\bf GE} denote the proposed $local\ encoder$ and $global\ encoder$, respectively.}
            \label{joint_process}
        \end{figure*}

\section{ Zero Pronoun-specific Neural Network}
In this section, we describe the proposed zero pronoun-specific neural network ({\bf ZPSNN}) for AZP resolution. We first give an overview of the approach. Afterwards, we present each part of the {\bf ZPSNN} in details. Lastly, we describe the training and initialization processes of our model.

\subsection{An Overview of the Approach}
Given an AZP $zp$, we first extract a set of candidate antecedents. Following \newcite{chen2015}, to avoid having to deal with a potentially large number of candidates, we consider all and only those NPs that are two sentences away at most from the given AZP to be its potential candidate antecedents. In addition, among these NPs, we qualify those who are either maximal NPs or modifier NPs as candidates. 

An illustration of {\bf ZPSNN} is given in Figure~\ref{joint_process}. Our approach consists of two parts, i.e., one for AZP and another for candidate antecedents. Suppose $k$ NP are extracted as candidate antecedents, the {\bf ZPSNN} aims at determining the correct antecedent of $zp$ from its candidate set $NP = \{np_1,np_2, ..., np_k\}$. In order to capture the semantics of words, we represent each word as a low dimensional, continuous and real-valued vector, also known as word embedding \cite{bengio2003}. All the word vectors are stacked in a word embedding matrix $L_w\in\mathbb{R}^{d\times|V|}$, where $d$ is the dimension of word vector and $|V|$ is the size of word vocabulary. The word embedding of word $w$ is notated as $e\in\mathbb{R}^{d\times1}$, which is the column in the embedding matrix $L_w$.

In an attempt to represent AZPs, we develop the ZP-center LSTM network that generates the representation of an AZP by utilizing its contextual information. We denote the encoder function as $f(zp)$ that produces the vector representation of $zp$. For the candidate antecedent part, we propose a two-level candidate antecedent encoder, namely the $candidate\ encoder$ that is composed of two distinct sub-networks, i.e., the $local\ encoder$ and the $global\ encoder$. First, the $local\ encoder$ models every candidate antecedent into its local vector representation by utilizing the content and contextual information of a single candidate. We then organize all these local vector representations as the inputs of the $global\ encoder$, which generates the global representations for candidate antecedents with respect to the whole candidate set. For convenience, we denote the local representation of the $i$-th candidate antecedent $np_i$ as $l(np_i)$ and its corresponding global representation in $NP$ as $g_i(NP)$. Finally, the {\bf ZPSNN} scores each candidate antecedent by passing their local and global representations together with the vector representation of AZP through a $tanh$ layer, which adaptively selects correct antecedent from the candidate set. Specifically, given $zp$ and its $i$-th candidate antecedent $np_i$, we compute the resolution score as $res(f(zp),l(np_i),g_i(NP))$. To form a proper probability distribution over the candidate antecedents in the candidates set, we normalize the scores using the $softmax$ function. This way, we compute the probability the $i$-th candidate antecedent, $np_i$, is the correct antecedent of $zp$ as:
        \begin{equation}
            P(np_i|zp,NP) \propto exp(res(f(zp),l(np_i),g_i(NP)))
        \end{equation}

\subsection{ZP-centered LSTM}
In an attempt to represent AZPs by their contextual information, for each AZP, a sequence of words from the beginning to the end of the sentence it appears are extracted as the inputs of AZP part. One way to encode the sequence of words is via a recurrent neural network (RNN)~\cite{elman1991distributed}, which has been widely exploited to deal with variable-length sequence input. LSTM \cite{hochreiter1997long} is one of the classical variations of RNN, aiming to facilitate the training of RNN by solving the diminishing and exploding gradient problems in the deep or long structures.

Intuitively, the words near an AZP generally contain more information to express it. With this motivation, on the basis of the traditional LSTM, we propose the ZP-centered LSTM to encode the AZP, as is shown in Figure~\ref{joint_process}. The basic idea is to model the preceding and following contexts of the AZP separately, so that words surrounding the AZP could be better utilized to represent the AZP.

Specifically, we employ two LSTM neural networks, namely, LSTM$_{pre}$ and LSTM$_{fol}$ to model the preceding and following contexts of the AZP, respectively. The input of LSTM$_{pre}$ is the preceding context of an AZP, and the input of LSTM$_{fol}$ is the following context. We run LSTM$_{pre}$ from left to right, and run LSTM$_{fol}$ from right to left. We favor this strategy as we believe that regarding the words near the AZP as the last hidden units could contribute more in representing the AZP and it also makes our model capable of exploiting long-distance context information of the AZP. Afterwards, we concatenate the last hidden vectors of LSTM$_{pre}$ and LSTM$_{fol}$ together to generate the representation of AZP. One could also try averaging or summing the last hidden vectors of LSTM$_{pre}$ and LSTM$_{fol}$ as alternatives.

\subsection{Modeling Candidate Antecedents}
We describe our method of generating representations of candidate antecedents in this subsection. As previous mentioned, we employ the two-level $candidate\ encoder$ that is composed of two sub-networks, i.e., the $local\ encoder$ that generates the local representation of a candidate antecedent by utilizing its content and contextual information; and the $global\ encoder$ that produces the global representation of a candidate antecedent by considering the other candidate antecedents in the whole candidate set.

For $local\ encoder$, its input contains the content and context words of a given candidate antecedent. In order to model both the inside and outside information of the phrase, we adopt the similar method utilized in \cite{clark2016improving} to generate the local representation of a candidate antecedent. In particular, various words and groups of words are extracted and fed into the $local\ encoder$, i.e., word embeddings of the head word, first word, last word, two preceding words, two following words; and averaged word embeddings of the five preceding context words, five following context words, all the contents words. We then concatenate these embeddings together and pass them through three hidden layers of rectified linear units (ReLu) \cite{nair2010rectified}. Afterwards, the output of the last hidden layer is regarded as the local representation for the given candidate antecedent. By doing so, $local\ encoder$ produces the local representations of all the candidate antecedents in the candidate set as $\{l(np_1), l(np_2), ..., l(np_k)\}$.

In addition, for purpose of capturing the global information of candidate antecedents, we develop the $global\ encoder$, which generates the vector representation of a given candidate antecedent from the global perspective by modeling all the candidate antecedents in the candidate set. To achieve this goal, a desirable solution should be capable of explicitly capturing useful information provided by the other candidate antecedents in the candidate set when dealing with one specific candidate antecedent. Admittedly, a helpful property of LSTM is that it could keep useful history information in the memory cell by exploiting input, output and forget gates to decide how to utilize and update the memory of previous information. Therefore, we develop the $global\ encoder$ by employing the LSTM neural network, which takes the outputs of the $local\ encoder$, $\{l(np_1), l(np_2), ..., l(np_k)\}$ as inputs. Specifically, our $global\ encoder$ is implemented as a bidirectional LSTM whose hidden states form the global representations of candidate antecedents, that is $g_i(NP) = \overrightarrow{g}_i(NP) || \overleftarrow{g}_i(NP)$ where $||$ denotes vector concatenation, $\overrightarrow{g}_i$ and $\overleftarrow{g}_i$ present forward and backward global representations from the respective recurrent networks. For the $i$-th candidate antecedent $np_i$, its corresponding hidden vector captures useful information of candidate antecedents before (after) and including $np_i$ from the global perspective.

In this manner, our two-level $candidate\ encoder$ generates both the local and global representations of candidate antecedents. We then feed these representations to the next layer to go through the remaining procedures of our system.

\subsection{Final Prediction}
After generating the representations of an AZP and its candidate antecedents, we predict the correct antecedents for the AZP by utilizing these continuous distributed representations. Specifically, we employ a $tanh$ layer to adaptively score the candidate antecedents. Moreover, as is shown in \newcite{chen2016}, traditional hand-crafted features are crucial for the resolver's success since they capture the syntactic, positional and other relationships between an AZP and its candidate antecedents. Following \newcite{chen2016} who encode hand-crafted features as inputs to their neural network model, we integrate a set of features that are employed in \cite{chen2016}, in the form of vector ($v^{(fe)}$) into our model.

In particular, taking the representation of an AZP $f(zp)$, its $i$-th candidate antecedent $np_i$ with the local representation $l(np_i)$ and global representation $g_i(NP)$, and the corresponding feature vector $v_{i}^{(fe)}$ as inputs, we compute the initial resolution score $s_i$ as:
\begin{equation}
\begin{split}
        s_i = tanh(W^{(s)}[f(zp);l(np_i);g_i(NP);v_{i}^{(fe)}] + b^{(s)})
\end{split}
\end{equation}
where $W^{(s)}$ and $b^{(s)}$ are parameters of the $tanh$ layer.
After obtaining $\{s_1,s_2,...,s_k\}$, we feed them to a $softmax$ layer to calculate the final resolution scores $\{score_1,score_2,...,score_k\}$ as:
        \begin{equation}
            score_i = \frac{exp(s_i)}{\sum_{i^{'}=1}^k exp(s_{i^{'}})}
        \end{equation}
Finally, we regard $score_i$ as the probability that the $i$-th candidate antecedent is the correct antecedent, and predict the highest-scoring (most probable) one to be the antecedent of the given AZP.

\subsection{Training and Initialization}
We train the {\bf ZPSNN} in an end-to-end way in a supervised manner. The model is trained by minimizing the cross-entropy error of coreference classification, whose loss function is given below:
      \begin{equation}
            loss = -\sum_{t \in T}\sum_{np \in NP}\delta(zp,np)\log(P(zp,np))
    \end{equation}
where $T$ represents all training instances, $NP$ is the candidate set of AZP $zp$.
$\delta(zp,np)$ is an indicator function indicating whether $zp$ and its candidate antecedent $np$ are coreferent:

$$\delta(zp,np)=
\begin{cases}
$1,$& \text{if $zp$ and $np$ are coreferent}\\
$0,$& \text{otherwise}
\end{cases}$$

We take the derivative of loss function through back-propagation with respect to the whole set of parameters, and update parameters with stochastic gradient descent by setting the learning rate as $0.01$. We use a pre-trained {100-dimensional} Chinese word embeddings\footnote{Embeddings are trained by {\bf word2vec} toolkit on Chinese portion of the training data from the OntoNotes 5.0 corpus.} as inputs. For parameters, we randomly initialize them from a uniform distribution $U(-0.01,0.01)$.

\section{Experiments}
\subsection{Corpus}
    We employ the dataset used in the official CoNLL-2012 shared task, from the OntoNotes Release 5.0, to carry out our experiment. The CoNLL-2012 shared task dataset consists of three parts, i.e. a training set, a development set, and a test set. Table~\ref{files} shows the statistics of our corpus.
        \begin{table}[h]\small
        \begin{center}
        \begin{tabular}{|c||c|c|c|c|}
        \hline
        & Documents & Sentences & ZPs & AZPs \\
        \hline
        Train & 1,391 & 36,487 & 23,065 & 12,111\\
        \hline
        Test & 172 & 6,083 & 3,658 & 1,713\\
        \hline
        \end{tabular}
        \end{center}
        \caption{\label{files} Statistics of our corpus.}
        \end{table}
        \\Documents in the corpus come from six sources, namely Broadcast News (BN), Newswires (NW), Broadcast Conversations (BC), Telephone Conversations (TC), Web Blogs (WB), and Magazines (MZ). Considering that only the training set and the development set are annotated with ZPs, we thus utilize the training set for training and the development set for testing. The same experimental data setting is utilized in our baseline system~\cite{chen2016}.

\newcommand{\PreserveBackslash}[1]{\let\temp=\\#1\let\\=\temp}
\newcolumntype{C}[1]{>{\PreserveBackslash\centering}p{#1}}
\newcolumntype{R}[1]{>{\PreserveBackslash\raggedleft}p{#1}}
\newcolumntype{L}[1]{>{\PreserveBackslash\raggedright}p{#1}}

         \begin{table*}
           \small
            \begin{center}
            \begin{tabular}{|c|| C{0.35cm} C{0.35cm} C{0.45cm}| C{0.35cm} C{0.35cm} C{0.45cm}|| C{0.35cm} C{0.35cm} C{0.45cm}| C{0.35cm} C{0.35cm} C{0.45cm}|| C{0.35cm} C{0.35cm} C{0.45cm}|C{0.35cm} C{0.35cm} C{0.45cm}|}
            \hline
            \multicolumn{1}{|c||}{} & \multicolumn{6}{c||}{\scriptsize {Setting 1: Gold Parse + Gold AZP}} & \multicolumn{6}{c||}{\scriptsize {Setting 2: Gold Parse + System AZP}} & \multicolumn{6}{c|}{\scriptsize {Setting 3: System Parse + System AZP}}\\
            \cline{2-19}
            \multicolumn{1}{|c||}{} & \multicolumn{3}{c}{Baseline} & \multicolumn{3}{|c||}{{\bf ZPSNN}} & \multicolumn{3}{c}{Baseline} & \multicolumn{3}{|c||}{{\bf ZPSNN}} & \multicolumn{3}{c}{Baseline} & \multicolumn{3}{|c|}{{\bf ZPSNN}$^\dagger_*$}\\
            \cline{2-19}
            \multicolumn{1}{|c||}{} &R&P&F&R& P & F & R & P & F & R & P & F & R & P & F & R & P & F\\
            \hline
                Overall & 51.8 & 52.5 & 52.2 & 53.4 & 53.8 & {\bf 53.6} & 39.6 & 27.0 & 32.1 & 39.7 & 29.7 & {\bf 34.0} & 21.9 & 15.8 & 18.4 & 22.1 & 19.9 & {\bf 20.9} \\
            \hline\hline
                NW & 48.8 & 48.8 & 48.8 & 50.0 & 50.0 & {\bf 50.0} & 34.5 & 26.4 & 29.9 & 29.7 & 34.2 & {\bf 31.8} & 11.9 & 12.8 & 12.3 & 21.4 & 21.2 & {\bf 21.3}\\
            \rowcolor{mygray}
                MZ & 41.4 & 41.6 & 41.5 & 45.0 & 45.0 & {\bf 45.0} & 34.0 & 22.4 & 27.0 & 30.8 & 29.9 & {\bf 30.4} & 9.3 & 7.3 & 8.2 & 13.6 & 13.3 & {\bf 13.5}\\
                WB & 56.3 & 56.3 & {\bf 56.3} & 55.6 & 56.2 & 55.9 & 44.7 & 25.1 & {\bf 32.2} & 38.7 & 25.6 & 30.9 & 23.9 & 16.1 & 19.2 & 23.2 & 19.7 & {\bf 21.3}\\
            \rowcolor{mygray}
                BN & 55.4 & 55.4 & {\bf 55.4} & 53.3 & 53.3 & 53.3 & 36.9 & 31.9 & {\bf 34.2} & 41.0 & 26.6 & 32.3 & 22.1 & 23.2 & 22.6 & 23.1 & 24.0 & {\bf 23.5} \\
                BC & 50.4 & 51.3 & 50.8 & 55.0 & 55.6 & {\bf 55.3} & 37.6 & 25.6 & 30.5 & 41.9 & 31.1 & {\bf 35.7} & 21.2 & 14.6 & 17.3 & 22.2 & 19.3 & {\bf 20.6} \\
            \rowcolor{mygray}
                TC & 51.9 & 54.2 & 53.1 & 53.7 & 55.1 & {\bf 54.4} & 46.3 & 29.0 & 35.6 & 41.3 & 36.3 & {\bf 38.7} & 31.4 & 15.9 & 21.1 & 24.7 & 20.9 & {\bf 22.7}\\
            \hline 
            \end{tabular}
            \end{center}
            \caption{\label{result}  Experimental results on the test data. {\bf ZPSNN} represents our approach. $^\dagger_*$ indicate that our approach is statistical significant over the baseline system (using t-test, with $p<0.05$). } 
            \end{table*}
\subsection{Evaluation Metrics}
    Following researches on Chinese zero pronoun resolution \cite{zhao2007,chen2016}, we evaluate the results in terms of recall (R), precision (P), and F-score (F). 
\subsection{Experimental Results}

We employ~\newcite{chen2016}'s system as the baseline, which is the state-of-the-art Chinese ZP resolution system. To evaluate our proposed approach, following~\newcite{chen2016}, three experimental settings are designed.  In Setting 1, we assume that gold AZPs and gold syntactic parse trees are available (obtained from the CoNLL-2012 shared task dataset).
In Setting 2, we utilize gold syntactic parse trees and system AZPs\footnote{In this study, we adopt the learning-based AZP identification approach utilized in \cite{chen2016} to identifiy system AZPs.}. Finally, in Setting 3, we employ system AZP and system syntactic parse trees that obtained by Berkeley parser, which is the state-of-the-art parsing model. Experimental results of our approach and the baseline system are shown in Table~\ref{result}. The first row in Table~\ref{result} is the overall scores, follows are results in different sources in test data. As we can see, our approach significantly outperforms the baseline system under three experimental settings by 1.4\%, 1.9\%, 2.5\% in terms of overall F-score, respectively. 
In per-source results, for Setting 1 and 2, {\bf ZPSNN} beats the baseline system in four of the six sources of data, especial on the source BC, which has the most AZPs in six sources. Moreover, in Setting 3, our system achieves higher performance than the baseline system in all the six sources of data. All these approve that our proposed approach achieves a considerable improvement in Chinese ZP resolution. 

Moreover, to demonstrate the utility of global and local information of candidate antecedents, we conduct a set of ablation experiments, as shown in Table~\ref{tuning1}. In each ablation experiment, we retrain the model with only one type of representations of candidates utilized as inputs. To minimize the external influence, we simply employ experimental Setting 1 (gold parse and gold AZP) to perform the experiments. We can observe that incorporating the global information significantly boosts the F-score comparing with the model utilize only the local information. Meanwhile, local representations of candidate antecedents provide the system with a strong ability to reveal the semantics of the phrases. Consequently, when the local information is ablated (i.e., without utilizing $l(np_i)$ when computing the formula 2), F-score drops by 0.7\% comparing with the full system.

       \begin{table}[htb!]\small
       \begin{center}
      \begin{tabular}{|c||c|c|c|}
    \hline
        & R & P & F\\
        \hline
        Full system & {\bf 53.4} & {\bf 53.8} & {\bf 53.6} \\
        Global information only & 52.7 & 53.1 & 52.9 \\
        Local information only & 51.6 & 52.0 & 51.8 \\
        \hline
      \end{tabular}
        \end{center}
        \caption{\label{tuning1} Ablation experimental results on the whole test data.}
        \end{table}

\begin{figure}[h]
            \centering
            \includegraphics[width=0.4\textwidth]{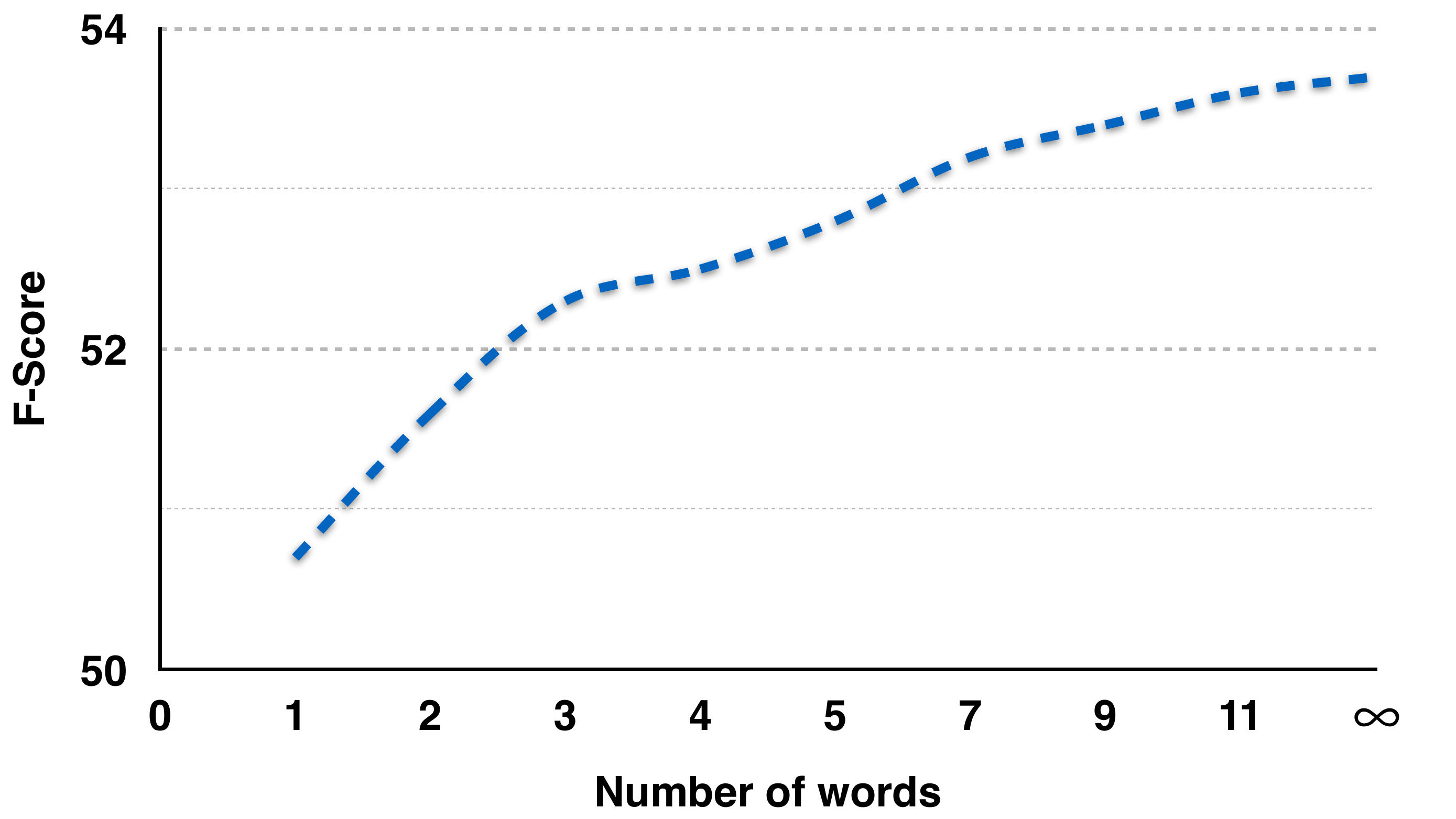}
            \caption{Effect of tuning the number of context words extracted for modeling the AZP, where $\infty$ means extracting all the words from the beginning to the end of the sentence.}
            \label{tuning}
            \end{figure}

Recall that in our model, we extract a sequence of context words to represent an AZP. Obviously, it is crucial to our model that how many context words should be encoded to describe the AZP. Hence we conduct experiments by tuning the number of context words we extracted for encoding the AZP. To minimum the external influence, we employ experimental Setting 1 to carry out the experiment. Figure~\ref{tuning} shows the experimental results, where the number ``x'' in x-axis means extracting $x$ nearest words from both an AZP's preceding and following contexts.

From Figure~\ref{tuning} we can observe that the performance of our model is improved by introducing more context words. Therefore, we encode an AZP by utilizing all its preceding and following context words in our system. Such a phenomenon also indicates the effects of contextual information we utilized to represent the AZP. 

\subsection{Error Analysis}

To better evaluate our proposed approach, we perform a case study for experimental results. Our analysis reveals that there are mainly three types of errors, as discussed below.
    
First, our model may fail when words in candidate antecedents are not in embedding matrix.
        \begin{quote}\small
        威迪奥诺 \ 表示 \ {\bf$\phi$} \ 将 \ 把 \ 意见 \ 转达 \ 给 \ 联合国秘书长 。\\
        (Weidiaunuo expressed that {\bf $\phi$ } will convey this suggestion to the UN secretary general.)
        \end{quote}
In this case, the correct antecedent of AZP ``$\phi$'' is NP ``威迪奥诺/Weidiaunuo'' while our model predicts NP ``国家/the country'' (appears in the preceding contexts) as the result, which is incorrect. We observe that the word ``威迪奥诺/Weidiaunuo'' has an embedding as $Null$, which makes our model incapable of representing the candidate antecedent, draws incorrect results. Hence, some more efforts can be achieved, as we can present words that are not in our embedding matrix with appropriate expressions.
 
The second type of error appears when candidate antecedents contain lots of content words, such as ``英国的查尔斯王子和他的夫人卡米拉/Britain's prince William and his wife Camilla''. When modeling a candidate antecedent, our method encodes its local representation by utilizing groups of its content and context words. In particular, we employ the average word embedding as the content expression of a candidate antecedent. However, such an averaging operation treats all the words in a phrase equally, which causes difficulty in well handling complex expression of candidate antecedents consisting of many words. One way to address this problem would be to explicitly focus on the more informative part of the phrase, for instance, applying some attention-based neural network models when representing the candidate antecedents.

The third type of error appears when an AZP lies at the beginning of a sentence, our model prefers to choose pronoun NPs like ``你/you'' or ``我/I'', as antecedents.
    \begin{quote}\small
            {\bf $\phi$ }\ 为何\  成为 \ 羡慕 \ 且 \ 嫉妒 \ 的 \ 目标？\\
            (Why {\bf $\phi$ } becomes the focus of envy and jealousy?)
    \end{quote}
In this case, ZP ``$\phi$'' should be resolved to NP ``台北市/Taibei city'' while our model chooses the NP ``我们/we''. As many of the sentences in our training data start with an overt pronoun, our model thus prefers pronouns to fill the AZP gap. In addition, the {\bf ZPSNN} tries to encode an AZP by utilizing the words appear both in its preceding and following contexts. Unfortunately, for aforementioned AZPs, there are no preceding context words. The problem can be better alleviated if we take more contexts into account, by capturing word sequences across the sentence boundaries, acquiring sentence-level history vectors, which will be exploited by us in future work.

\section{Related Work}
    
    \subsection{Chinese Zero Pronoun Resolution}
    Early studies employed heuristic rules to Chinese ZP resolution. \newcite{converse2006} proposes a rule-based method to resolve the zero pronouns, by utilizing Hobbs algorithm \cite{hobbs1978resolving} in the CTB documents.  More recently, supervised approaches to this task have been vastly explored. \newcite{zhao2007} present a supervised machine learning approach to the identification and resolution of Chinese zero pronouns. By employing the J48 decision tree model, several kinds of features are integrated into the resolution algorithm.  \newcite{kong2010} develop a novel approach for Chinese ZP resolution, employing context-sensitive convolution tree kernels to capture syntactic information. \newcite{chen2013} further extend \newcite{zhao2007}'s study by considering more types of novel features. Moreover, they exploit co-reference links between zero pronouns and antecedents. In recent time, \newcite{chen2014} develop an unsupervised language-independent approach. They first recover each zero pronouns into ten overt  pronouns and then apply a ranking model to rank the candidate antecedents. \newcite{chen2015} propose an end-to-end unsupervised probabilistic model for Chinese ZP resolution, using a salience model to capture discourse information. \newcite{chen2016} propose a deep neural network approach to learn useful task-specific representations and effectively exploit lexical features via word embeddings. Experimental results on the Chinese portion of the OntoNotes 5.0 corpus show that their approach achieves the state-of-the-art performance. 

    \subsection{Zero Pronoun Resolution for Other Languages}
    There have been many researches on ZP resolution for other languages. These studies can be divided into rule-based and supervised machine learning approaches. \newcite{ferrandez2000} propose a set of hand-crafted rules for Spanish ZP resolution. Recently, supervised approaches have been exploited for ZP resolution in Korean \cite{han2006korean} and Japanese \cite{isozaki2003japanese,iida2006,iida2007zero,imamura2009discriminative,sasano2011,iida2011cross}. \newcite{iida2015intra} propose a novel approach of recognizing subject sharing relations, where an ILP-based zero anaphora resolution method is combined to improve the performance of intra-sentential zero anaphora resolution in Japanese.

\section{Conclusion}
    In this study, we investigate a novel zero pronoun-specific neural network ({\bf ZPSNN}) for Chinese zero pronoun resolution, exploiting a learning algorithm that is capable of modeling contextual information to represent zero pronouns at the semantic level. Besides, a two-level candidate antecedent encoder is employed to explicitly capture the global and local information of candidate antecedents with respect to the whole candidate set. Experimental results on the Chinese portion of the OntoNotes 5.0 corpus show that our proposed approach outperforms the state-of-the-art methods in various experimental settings.

\section*{Acknowledgments}
We greatly thank Xuxiang Wang for tremendously helpful discussions. We also thank the anonymous reviewers for their valuable comments. This work was supported by the National High Technology Development 863 Program of China (No.2015AA015407), National Natural Science Foundation of China (No.61472105 and No.61472107).

\bibliographystyle{named}
\bibliography{reference}

\end{CJK*}
\end{document}